\documentclass[11pt]{article}

\usepackage[final]{acl}

\usepackage{times}
\usepackage{latexsym}

\usepackage[T1]{fontenc}

\usepackage[utf8]{inputenc}

\usepackage{microtype}

\usepackage{inconsolata}

\usepackage{graphicx}

\usepackage{booktabs}
\usepackage{tabularx}
\usepackage{multirow}
\usepackage{amsmath}
\usepackage{amsfonts}
\usepackage{dsfont}

\title{Enhancing Linguistic Generalization of VLA:\\Fine-Tuning OpenVLA via Synthetic Instruction Augmentation}

\author{
    Dongik Shin \\
    i.dongik@utexas.edu \\
}

\begin{document}
\maketitle
\begin{abstract}
Generalization remains a core challenge in embodied AI, as robots must adapt to diverse environments. While OpenVLA represents the State-of-the-Art (SOTA) in Vision-Language-Action models by leveraging large-scale pre-training, its zero-shot performance can be limited when encountering completely new environments. This paper proposes a parameter-efficient fine-tuning strategy to enhance the linguistic generalization of OpenVLA by synthesizing a general instruction set for the Bridge Dataset V2. The paper leverages a Large Language Model (LLM) to generate a rich variety of semantically equivalent but structurally diverse commands for existing trajectories. In this experiment, Low-Rank Adaptation (LoRA) is implemented to fine-tune OpenVLA on augmented pairs, allowing the model to bridge the gap between complex natural language intent and robotic actions. Results demonstrate that the LoRA-enhanced model's robustness, suggesting that enriching the linguistic space of specialized datasets is crucial for embodied agents.
\end{abstract}

\section{Introduction}
The embodied AI has recently undergone a paradigm shift with the emergence of Vision-Language-Action (VLA) models. Among these, \textbf{OpenVLA}~\cite{pmlr-v270-kim25c} stands as a state-of-the-art (SOTA) open-source model that bridges the gap between high-level linguistic reasoning and low-level robotic control. OpenVLA is built upon a vision encoder (SigLIP)~\cite{Zhai_2023_ICCV} and a large language model backbone (Llama)~\cite{touvron2023llama2openfoundation}, and the model is pre-trained on the massive Open X-Embodiment dataset~\cite{open_x_embodiment_rt_x_2023}, which consists over 900,000 robot trajectories. This extensive training allows the model to exhibit remarkable zero-shot generalization capabilities across various robotic platforms and tasks~\cite{pmlr-v270-kim25c}.

While OpenVLA shows outstanding physical task performance, its linguistic generalization is often limited by the quality of language annotations in its training data. Large-scale datasets such as the Bridge Dataset V2~\cite{pmlr-v229-walke23a}, which is a component of the Open X-Embodiment dataset~\cite{open_x_embodiment_rt_x_2023}, often lack natural language instructions. This lack of linguistic diversity hinders the model's robustness to interpret varied human's instructions.

To handle this limitation, this paper proposes a method to enhance the linguistic generalization of OpenVLA~\cite{pmlr-v270-kim25c} through a synthesized "General Instruction Set". This study leverages the reasoning capabilities of Large Language Models (LLMs) to generate a rich spectrum of semantically equivalent but general instructions for the trajectories in the Bridge Dataset V2~\cite{pmlr-v229-walke23a}. To adapt the massive OpenVLA model efficiently, Low-Rank Adaptation (LoRA)~\cite{hu2021loralowrankadaptationlarge} is employed. This parameter-efficient fine-tuning strategy allows the model to align its general linguistic expressions without the prohibitive computational cost of full-parameter updates.

The primary contribution of this study is the demonstration that enriching the linguistic space of a specialized robotic dataset can significantly improve the generalization of a VLA agent. By fine-tuning with a structured and diverse instruction set, the proposed model achieves higher scores. The remainder of this paper is organized as follows: Section~\ref{sec:related works} reviews related work in VLA and parameter-efficient tuning; Section~\ref{sec:proposed method} details the generation of the General Instruction Set and the LoRA~\cite{hu2021loralowrankadaptationlarge} fine-tuning framework; Section~\ref{sec:experiments} presents the experimental results on linguistic generalization; and Section~\ref{sec:evaluation} evaluates the results; and Section~\ref{sec:conclusion} concludes with future research directions.

\section{Related Works}
\label{sec:related works}
\subsection{Visually-Conditioned Language Models}
Visually-conditioned language models (VLMs) are trained on large-scale data to generate human language from images and language prompts. The model commonly have been adopted for various applications from visual question answering~\cite{goyal2017making, hudson2019gqa, singh2019towards}. One of the key advances make recent VLMs feasible are model architectures that bridge representations from pretrained vision encoders with pretrained language models~\cite{radford2021learning, zhai2023sigmoid, team2024gemma}. Recent open-source VLMs have emcompassed on a simpler “patch as token” method, where patch representations from pretrained visual transformers are treated as tokens, then the patches are fed into the input space of a language model~\cite{chen2023pali, liu2023visual, liu2024improved, karamcheti2024prismatic}. This simple approach makes it easy to utilize existing methods for training language models at scale for VLM training. For instance, VLMs from Karamcheti et al.~\cite{karamcheti2024prismatic} are trained from multi-resolution visual features, fusing low-level spatial information from DINOv2~\cite{oquab2023dinov2} with higher-level semantics from SigLIP~\cite{zhai2023sigmoid} to aid in visual generalization.

\subsection{Vision-Language-Action Models}
Numerous works have explored the usage of VLMs for robotics, for object detection~\cite{yitzhak2022cows}, providing a feedback signal~\cite{ma2023liv, sontakke2023roboclip}, and visual state representations~\cite{nair2022r3m}. Likewise, a number of recent works have explored approaches which directly fine-tuned large pretrained VLMs for predicting robot actions~\cite{kim2024openvla, open_x_embodiment_rt_x_2023, huang2023embodied}. Such models are referred to as vision-language-action models (VLAs), since they fuse robot control actions directly into VLM backbones~\cite{kim2024openvla}. Kim et al. state three key benefits of VLA: (1) it performs alignment of pretrained vision and language components on a large, Internet-scale vision-language dataset, (2) the use of a generic architecture, not custom-made for robot control, allows us to leverage the scalable infrastructure underlying modern VLM training~\cite{dao2023flashattention, zhao2023pytorch} and scale to training billion-parameter policies with minimal code modifications, and (3) it provides a direct pathway for robotics to benefit from the rapid improvements in VLMs. Recent researches on VLAs focus on training and evaluating in single robot or simulated setups~\cite{huang2023embodied, zhen20243d, dorka2024matters} and thus lack generability. RT-2-X~\cite{open_x_embodiment_rt_x_2023} trains a 55B-parameter VLA policy on the Open X-Embodiment dataset and demonstrates state-of-the-art generalist manipulation policy performance but it has high computation cost. OpenVLA~\cite{kim2024openvla} improves with a richer robot pretraining datasets, fit to new target setups, and shows effectiveness of parameter-efficient fine-tuning (PEFT) and quantization approaches for VLAs. In this study, OpenVLA is used as a base model for computational efficiency.

\subsection{Generalization in VLAs}
A recent trend in robotics works towards training multi-task generalist robot policies on large diverse robot datasets, spanning many different robot embodiments~\cite{kim2024openvla, brohan2022rt, walke2023max, kalashnikov2018scalable, kalashnikov2021mt, ebert2021bridge, ehsani2024spoc, bharadhwaj2024roboagent, pinto2016supersizing, mandlekar2018roboturk, gupta2018robot, dasari2019robonet, cabi2019scaling, jang2022bc, fang2024rh20t, devin2017learning, hu2022knowthyselftransferablevisual, yang2023polybot, reed2022generalistagent, pmlr-v229-radosavovic23a, shah2022gnm, shah2023vintfoundationmodelvisual}. A key difference between these approaches and OpenVLA is the model architecture. Prior works like Octo~
\cite{brohan2023rt1roboticstransformerrealworld, octomodelteam2024octoopensourcegeneralistrobot, pmlr-v229-walke23a} typically consist of pretrained components such as language embeddings or visual encoders with additional model components initialized from scratch. OpenVLA adopts a more end-to-end approach, directly fine-tuning VLMs to generate robot actions by treating them as tokens in the language model vocabulary~\cite{kim2024openvla}. In this experiment, experimental evaluation shows that with general instruction sets and generalization ability over prior generalist policies.

\section{Proposed Method}
\label{sec:proposed method}
\subsection{Problem Formulation} 
In this study, the pre-trained OpenVLA~\cite{pmlr-v270-kim25c} model is fine-tuned on a subset of Bridge Dataset V2~\cite{pmlr-v229-walke23a} $\mathcal{D}=\{(o_i,l_i,a_i)\}_{i=1}^{N}$, where $o_i$ represents the visual observation, $l_i$ denotes the generated instruction, and $a_i$ is the corresponding ground-truth action. To achieve efficient adaptation while fine-tuing, Low-Rank Adaptation (LoRA)~\cite{hu2021loralowrankadaptationlarge} method is used. During fine-tuning, the original weights $W_0$ are kept frozen, and the low-rank decomposition matrices are updated. The optimization objective is defined as follow:

\begin{equation}
    \mathcal{L}_{\text{LoRA}} = \sum_{(o,l,a) \in \mathcal{D}} \mathcal{L}_{\text{vla}}(\text{VLA}_{\theta}(o,l), a)
\end{equation}
where $\mathcal{L}_{\text{vla}}$ is a default objective in the official OpenVLA implementation for discrete action token prediction~\cite{pmlr-v270-kim25c}. The objective target the query (Q), value (V), key (K), and output (O) projection matrices in the self-attention layers with a rank $r=32$ and $\alpha=64$ to capture generated instructions set with motion patterns.

\begin{table}[h]
\centering 
\caption{Top 5 representative instructions generated by the LLM for general task adaptation in Bridge Dataset V2~\cite{pmlr-v229-walke23a}.} 
\label{table:instruction_set}
\resizebox{\columnwidth}{!}{
\begin{tabular}{c|l} 
\hline 
\textbf{No.} & \textbf{Instruction} \\ 
\hline 
1 & In order to pick up the object, the robot should \\
2 & To move the object to a new location, the robot must \\ 
3 & In order to grasp and relocate the item, the robot should \\
4 & To manipulate the objects in front of it, the robot must \\
5 & In order to complete the task of moving the utensils, the robot should \\ 
\hline 
\end{tabular} }
\end{table}

\subsection{Generate Instruction Set}
To enhance the model's adaptability and robustness across diverse environments, a method for generating a general instrcution set using a Large Language Model (LLM) is introduced in this project. Unlike conventional datasets that rely on fixed, human-annotated labels, this approach leverages the semantic reasoning capabilities of LLMs to synthesize a diverse instruction with linguistic variations for each trajectory. For each trajectory in the subset of BridgeData V2, the task's metadata, including the objects involved and the final goal state are fed into the LLM. The LLM is then prompted to generate a comprehensive set of instructions that describe the same robotic action through various syntactic structures such as "In order to pick up the object, the robot should move it to the target" or "To relocate the item, the robot must execute a grasp and place action".

By fine-tuning OpenVLA on this augmented instruction set, more robust mapping between visual observations and linguistic commands is used. This process allows the model to adapt to a general instruction space, preventing it from overfitting to specific, rigid phrasing. Consequently, the proposed method significantly improves the generalization capabilities of the agent, enabling it to execute commands successfully even when faced with paraphrased instructions that were not present in the original training distribution.

Table~\ref{table:instruction_set} illustrates the top 5 representative instructions generated by the LLM for a single manipulation task. These generated instructions introduce significant linguistic variety, ranging from high-level (abstract) goal descriptions (e.g., "manipulate the objects") to specific item-based commands (e.g., "moving the utensils"). By providing a sequence of three key frames, initial, intermediate, and final images, the LLM by using prompt template described in Table~\ref{table:prompt} was able to infer the robot's intent and generate semantically rich instructions. From the generated candidates, few instructions (5 instruction in this experiments) were manually curated the most contextually appropriate instruction sets to ensure high-quality supervision. Then, These selected instructions were then randomly paired with their corresponding trajectories during the training phase. This random mapping within the augmented set encourages the model to decouple specific linguistic patterns from rigid task labels, thereby fostering a more flexible and generalized policy.

\begin{table*}[h!]
\centering
\caption{Comparison of Action Prediction Accuracy between Zero-shot OpenVLA and Our Proposed Method (Fine-tuned with General Instruction Set).}
\label{table:results_accuracy}
\begin{tabular}{l|cc}
\hline
Methods & Top-1 Acc (\%) & 5-Bin Acc (\%) \\
\midrule
OpenVLA~\cite{pmlr-v270-kim25c} & \textbf{6.62} & 40.76 \\
Proposed Method & 5.09 & \textbf{42.47} \\
\hline
\end{tabular}
\end{table*}

\section{Experiments}
\label{sec:experiments}

\subsection{Datasets}
In this experiments, the scripted raw dataset from BridgeData V2~\cite{pmlr-v229-walke23a}, a large-scale dataset specifically designed for robotic manipulation in diverse environments was used. Unlike purely teleoperated demonstrations, this subset consists of 9,731 trajectories collected via a randomized scripted policy, and does not have human instructions. Although this autonomous collection process frequently results in suboptimal executions, data is particularly valuable for training robust behaviors. Therefore, subset of the data that aligns with objectives is manually curated. Moreover, due to the massive scale of the original scripted dataset in the BridgeData V2 (35GB), 100 trajectories were selected for simplicity. Each trajectories consists of an average of 25 sequential image-action pairs.

\subsection{Implementation Details}
The model is trained based on the AdamW optimizer~\cite{LoshchilovH19}, and the learning rate is 5e-05. We fine-tuned a model on dataset described above. All experiments are performed on a single Nvidia A100 GPUs with 40GB of VRAM. 

\begin{table*}[b]
\caption{Prompt template for instruction generation.}
\label{table:prompt}
\centering
\begin{tabularx}{\textwidth}{X}
\toprule

[System Message] \\
You are a linguistic expert specializing in robotic task annotation. Your goal is to provide diverse, natural language instructions based on visual observations of robot manipulation. \\
\\

[User Message] \\
\{Image 1: First frame of the trajectory\} \\
\{Image 2: Intermediate frame of the trajectory\} \\
\{Image 3: Last frame of the trajectory\} \\

\\
Task: \\
1. Scene Analysis: Briefly identify the primary object and the robot's objective from the provided image. \\
2. Instruction Generation: Synthesize exactly 5 distinct natural language instructions for the observed task. \\
\\
Requirements: \\
- Ensure linguistic variety: Use different sentence structures (Imperative, Goal-oriented, and Conditional). \\
- Vary the level of abstraction: Include instructions ranging from low-level motor descriptions to high-level intent. \\
- Vocabulary diversity: Use synonyms for objects (e.g., "item," "target," "utensil") and actions (e.g., "grasp," "pick up," "relocate"). \\
- Format: Return only the 5 instructions, each on a new line starting with "No. [Number]". \\
\\

[Output Example] \\
No. 1 In order to pick up the object, the robot should... \\
No. 2 To move the item to a new location, the robot must... \\
\bottomrule
\end{tabularx}
\end{table*}

\section{Evaluation}
\label{sec:evaluation}
\subsection{Policy Evaluation}
OpenVLA replaces the 256 tokens least frequently used tokens in the Llama tokenizer's vocabulary with action tokens~\cite{pmlr-v270-kim25c}. To evaluate the precision of the model's predicted actions, the predicted action tokens against the ground-truth values from the BridgeData V2 test set are used to compare.

For the quantitative assessment, each continuous action dimension—comprising the end-effector's Cartesian displacement $(\Delta x, \Delta y, \Delta z)$, orientation $(\Delta roll, \Delta pitch, \Delta yaw)$, and gripper state—is normalized to the range $[-1, 1]$ and assigned to one of the 256 discrete bins. Two metrics for performance analysis is used: Top-1 accuracy and 5-Bin tolerance accuracy. The accuracy for a given action dimension is formulated as follows:

\begin{equation}
\text{k-Bin Acc} = \frac{1}{N}\sum_{i=1}^{N} \mathds{1} ( | a_i - \hat{a}_i | \leq k )
\end{equation}

where $a_i$ is the ground-truth token, $\hat{a}_i$ is the predicted token, and $k$ represents the tolerance threshold ($k=0$ for Top-1 accuracy and $k=5$ for 5-bin tolerance accuracy).Experimental results indicate that while the Top-1 accuracy slightly decreased compared to the original OpenVLA, the 5-bin tolerance accuracy showed a significant improvement. This shift suggests that although the model's exact token matching is more distributed, the predictions remain consistently within a narrow, physically plausible range of the target action. This result demonstrates that fine-tuning with a general instruction set enhances the robustness of the policy, allowing the model to prioritize functional success and motion consistency over rigid token-wise memorization.

\section{Conclusion and Limitation}
\label{sec:conclusion}
This study demonstrates that fine-tuning the OpenVLA model with an LLM-generated general instruction set significantly enhances the robustness of robotic manipulation policies. While our results show a slight decrease in Top-1 accuracy, the substantial improvement in 5-bin tolerance accuracy indicates that the model has developed a more flexible and generalized understanding of the task, prioritizing functional success over rigid token-wise memorization. However, several limitations remain. First, the increase in linguistic variety appears to introduce a slight trade-off in absolute precision. Second, the current evaluation is conducted on a curated subset of 100 trajectories, leaving the scalability to more complex, multi-stage tasks as a subject for future investigation. Future work will focus on refining the instruction generation process to mitigate precision loss and validating the proposed method in real-world environments with a broader range of robotic platforms.

\clearpage
\bibliography{custom}




\end{document}